\documentclass[conference]{IEEEtran}
\IEEEoverridecommandlockouts
\usepackage{cite}
\usepackage{amsmath,amssymb,amsfonts}
\usepackage{algorithmic}
\usepackage{graphicx}
\usepackage{textcomp}
\usepackage{xcolor}
\usepackage{booktabs}
\usepackage{float}
\usepackage{threeparttable}
\usepackage{multirow} 
\usepackage{url}
 \usepackage{hyperref}
\def\BibTeX{{\rm B\kern-.05em{\sc i\kern-.025em b}\kern-.08em
    T\kern-.1667em\lower.7ex\hbox{E}\kern-.125emX}}

\begin{document}

\title{MeniOmni: A Structured Multimodal Benchmark for Holistic Meniscus Injury Assessment}
\author{
    \IEEEauthorblockN{
        Shurui Xu\textsuperscript{1,3,*}, 
        Siqi Yang\textsuperscript{2,*}, 
        Weiping Ding\textsuperscript{3,4,\textdagger}, 
        Hui Wang\textsuperscript{1}, 
        Mengzhen Fan\textsuperscript{5}, 
        Yuyu Sun\textsuperscript{6} and 
        Shuyan Li\textsuperscript{1,7, \textdagger}
        \thanks{* These authors contributed equally to this work.}
        \thanks{\textdagger Corresponding authors. Contact emails: dwp9988@163.com (W.~Ding), shuyan.li@qub.ac.uk (S.~Li)}
    }
    
    \IEEEauthorblockA{\textsuperscript{1}School of Electronics, Electrical Engineering and Computer Science, Queen's University Belfast, Belfast, UK}
    \IEEEauthorblockA{\textsuperscript{2}Radiology Department, Affiliated Nantong Clinical College of Nantong University, Nantong First People's Hospital,\\ School of Clinical Medicine, Nantong University, Nantong, Jiangsu, China}
    \IEEEauthorblockA{\textsuperscript{3}School of Artificial Intelligence and Computer Science, Nantong University, Nantong, China}
    \IEEEauthorblockA{\textsuperscript{4}Faculty of Data Science, City University of Macau, Macau, China}
    \IEEEauthorblockA{\textsuperscript{5}Department of Chemistry, University of Oxford, Oxford, UK}
    \IEEEauthorblockA{\textsuperscript{6}Orthopedics Department, Nantong First People's Hospital, Southeast University, Nantong, Jiangsu, China}
    \IEEEauthorblockA{\textsuperscript{7}Institution for Ocean Engineering, Tsinghua SIGS, Shenzhen, China}
}
\maketitle

\begin{abstract}
Clinical diagnosis of meniscus injuries requires radiologists to integrate volumetric MRI evidence with patient context (e.g., sex, age, BMI) and to produce structured diagnostic reports. Existing knee MRI benchmarks are typically unimodal and rely on coarse labels, limiting their ability to evaluate holistic clinical reasoning. We introduce MeniOmni, a structured multimodal benchmark for meniscus injury assessment, consisting of 746 multi-center MRI studies with tri-planar volumetric inputs, Clinical Priors, and expert-annotated clinical text. MeniOmni supports two tasks: (1) fine-grained Stoller severity grading and (2) diagnostic report generation. We further propose risk-aware ordinal evaluation and a semantic consistency metric (Meni-Score) to better reflect clinical relevance. Baseline experiments show that incorporating Clinical Priors improves grading performance and reduces severe errors, highlighting the value of multimodal context for safer assessment. Code and data are available at \url{https://github.com/ShuruiXu/MeniOmni}.
\end{abstract}


\section{Introduction}
\label{sec:intro}

The knee meniscus is a critical biomechanical stabilizer, and meniscal tears account for one-third of knee arthroscopies annually~\cite{R1,R2}. While MRI is the diagnostic gold standard~\cite{R3}, clinical interpretation goes beyond pixel-level perception: radiologists synthesize volumetric visual cues, integrate patient-specific Clinical Priors, and formulate semantic reports. The shortage of expert radiologists motivates automated systems capable of emulating this multimodal workflow~\cite{R4}.

\begin{table*}[t]
  \centering
  \small
  \caption{Functional comparison of MeniOmni against existing knee MRI benchmarks.}
  \label{tab:functional_compare}
  \setlength{\tabcolsep}{8pt} 
  \begin{threeparttable}
    \begin{tabular}{l|ccc|c|c|c}
      \toprule
      \multirow{2}{*}{\textbf{Benchmark}} & \multicolumn{3}{c|}{\textbf{Input Modalities}} & \textbf{Label} & \textbf{Supported} & \textbf{Core} \\
       & \textbf{Vision} & \textbf{Tabular} & \textbf{Text} & \textbf{Granularity} & \textbf{Tasks} & \textbf{Focus} \\
      \midrule
      MRNet~\cite{R5} & \checkmark & $\times$ & $\times$ & Binary (0/1) & Classification & Abnormality \\
      fastMRI~\cite{R7} & \checkmark & $\times$ & $\times$ & None & Reconstruction & Image Quality \\
      FastMRI+~\cite{R8} & \checkmark & $\times$ & $\times$ & Bounding Box & Detection & Localization \\
      KneeMRI~\cite{R9} & \checkmark & $\times$ & $\times$ & Slice-level & Segmentation & Pixel-wise \\
      MeniMV~\cite{R6} & \checkmark & $\times$ & $\times$ & 4-Grade & Grading & Visual Severity \\
      \midrule
      \textbf{MeniOmni (Ours)} & \textbf{\checkmark} & \textbf{\checkmark} & \textbf{\checkmark} & \textbf{4-Grade} & \textbf{Grading \& Gen.} & \textbf{Holistic Reasoning} \\
      \bottomrule
    \end{tabular}
  \end{threeparttable}
  \vspace{-0.4cm}
\end{table*}

However, progress toward multimodal diagnostic AI is hindered by two bottlenecks. In terms of data, existing public datasets are task-specific and limited to the visual modality. As detailed in Table~\ref{tab:functional_compare}, standard benchmarks such as MRNet~\cite{R5} rely on coarse binary labels and lack the fine-grained severity grading (0–3) that is essential for surgical planning. Although recent initiatives such as MeniMV~\cite{R6} have advanced the field with multi-view analysis, they remain restricted to unimodal visual inputs, systematically omitting structured patient context and diagnostic semantics. Equally critical, existing evaluation protocols rely on generic metrics (e.g., accuracy, n-gram overlap) that treat all errors equally, ignoring ordinal severity and rewarding fluency over factual correctness. The field thus lacks both the multimodal data and clinically aligned metrics necessary for reliable diagnostic agents.
\begin{figure}[t]
  \centering
   \includegraphics[width=\linewidth]{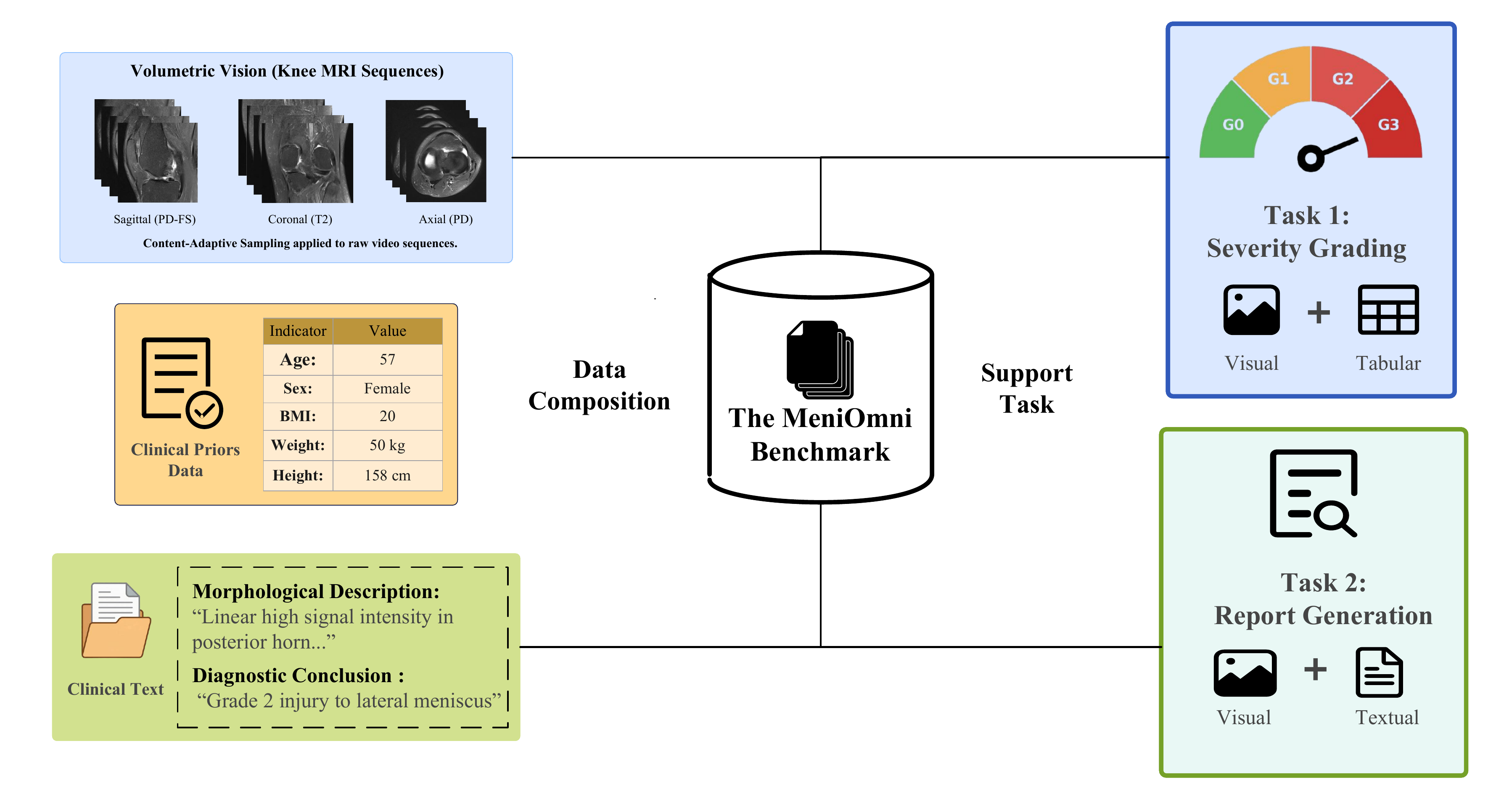} 
   \caption{Overview of the MeniOmni Benchmark construction and supported dual-task.}
   \label{fig:pipeline}
   \vspace{-0.5cm}
\end{figure}

To bridge this gap, we present MeniOmni, a structured multimodal benchmark supporting (1) Fine-Grained Severity Grading and (2) Diagnostic Report Generation. It combines three modalities: Volumetric Vision (content-adaptive tri-planar MRI sequences), Clinical Priors (tabular patient metadata), and Clinical Text (expert-annotated findings and conclusions).

Recognizing that generic evaluation metrics fail to capture the clinical validity of diagnostic reports and the ordinal nature of severity grading, we further introduce a set of task-specific evaluation functions. These metrics are meticulously designed to align model performance with clinical relevance, ensuring a rigorous assessment beyond superficial linguistic or classification accuracy. Extensive experimental results on MeniOmni validate that this multimodal integration is crucial for achieving both high accuracy and clinical safety.

The primary contributions of this work are:
\begin{itemize}
    \item We introduce \textbf{MeniOmni}, a multi-center benchmark that uniquely integrates volumetric MRI, structured Clinical Priors, and diagnostic text, addressing the scarcity of multimodal orthopedic datasets.
    
    \item We propose a Dual-Task Evaluation Protocol supporting both Fine-Grained Grading and Diagnostic Report Generation, enabling a rigorous assessment of MLLM capabilities.

    \item We establish comprehensive multimodal baselines by benchmarking state-of-the-art models on MeniOmni. This work defines the task landscape and quantifies the performance gain from multimodal integration.
\end{itemize}

\section{Related Work} 
\label{sec:related}

\subsection{Deep Learning for Knee MRI Analysis} 

Automated knee MRI analysis has evolved from 2D CNN segmentation~\cite{R5,R10} to 3D volumetric frameworks with attention mechanisms and Transformers~\cite{R11,R12}. However, these methods process MRI as a unimodal visual signal, neglecting structured patient metadata (age, BMI) that are clinically indispensable for distinguishing degeneration from acute tears~\cite{R1}.

\subsection{Multi-Modal Large Language Models in Medicine} 

Generalist biomedical MLLMs such as LLaVA-Med~\cite{R14} and RadFM~\cite{R15} have advanced radiology via instruction tuning but struggle in specialized orthopedics, failing to capture the spatiotemporal depth of 3D MRI and frequently hallucinating due to scarce expert-annotated musculoskeletal data~\cite{R16}.

\subsection{Benchmarks for Musculoskeletal Radiology} 

Existing musculoskeletal benchmarks such as MRNet~\cite{R5} and FastMRI+~\cite{R8} rely on coarse binary labels or detection tasks insufficient for fine-grained severity grading. Although MeniMV~\cite{R6} introduced ordinal grading (0--3), it remains confined to unimodal visual classification. MeniOmni bridges these gaps as the first benchmark integrating volumetric vision, tabular priors, and diagnostic text.

\section{The MeniOmni Benchmark}

\subsection{Data Acquisition and Standardization}

We present MeniOmni, the first structured multimodal benchmark specifically curated to support holistic orthopedic diagnostics. As illustrated in Figure~\ref{fig:pipeline}, the dataset is engineered to facilitate two complementary clinical tasks: (1) Fine-Grained Severity Grading (a multi-class classification problem) and (2) Diagnostic Report Generation (a multimodal text generation problem).

To emulate the complete cognitive workflow of radiologists, MeniOmni aggregates 746 cases from three clinical centers using three standard sequences (Sagittal PD-FS, Coronal T2, Axial PD). Each meniscus was independently graded by two musculoskeletal radiologists. The dataset is partitioned at the patient level 
in a 7:2:1 ratio (train/validation/test) to prevent data leakage. Each case is rigorously standardized into a triad of modalities:
\begin{itemize}
    \item \textbf{Volumetric Vision:} Content-adaptive MRI video sequences from three anatomical planes (Sagittal, Coronal, Axial).
    \item \textbf{Clinical Priors:} Key patient demographics (Age, Sex, BMI) that function as physiological risk factors.
    \item \textbf{Clinical Text:} Expert-annotated diagnostic reports comprising both morphological findings and conclusions.
\end{itemize}

Distinct from previous binary benchmarks, MeniOmni provides fine-grained severity supervision based on the Stoller grading system. Each meniscus (Medial and Lateral) is independently annotated on an ordinal scale: Grade 0 (Intact), Grade 1 \& 2 (Degeneration), and Grade 3 (Tear). This hierarchical schema ensures that models are evaluated not just on detecting abnormalities, but on quantifying pathological severity with clinical precision.

\subsection{Visual Modality: Content-Adaptive Volumetric Sampling}
To optimize visual inputs, we minimize redundancy within the MRI volume via a content-adaptive sampling algorithm. For a slice $I$, we quantify its value based on two dimensions: Image Quality ($S_{\text{IQ}}$) and Anatomical Complexity ($S_{\text{AC}}$). The scoring function is formulated as:

\begin{equation}
\label{eq:quality_score}
\begin{aligned}
S_{\text{IQ}} &= w_c \sigma(I) + w_s \text{Var}(\nabla^2 I) + w_e H(I), \\
S_{\text{AC}} &= \sum_{c \in \mathcal{C}} \ell(c) + \lambda \max_{c \in \mathcal{C}} \ell(c).
\end{aligned}
\end{equation}

\noindent where $\sigma(I)$ denotes the standard deviation representing local contrast; $\text{Var}(\nabla^2 I)$ is the variance of the Laplacian operator, capturing edge sharpness; and $H(I)$ represents Shannon entropy, quantifying overall information content. $w_c, w_s, w_e$ are hyperparameters balancing these factors. For anatomical complexity, $\ell(c)$ represents the confidence logits of meniscal regions detected by SAM-Med2D~\cite{R17}. By leveraging SAM-Med2D's promptable segmentation capabilities, we ensure the selected slices contain relevant anatomy rather than background noise.
\begin{figure}[t]
  \centering
   \includegraphics[width=1\linewidth]{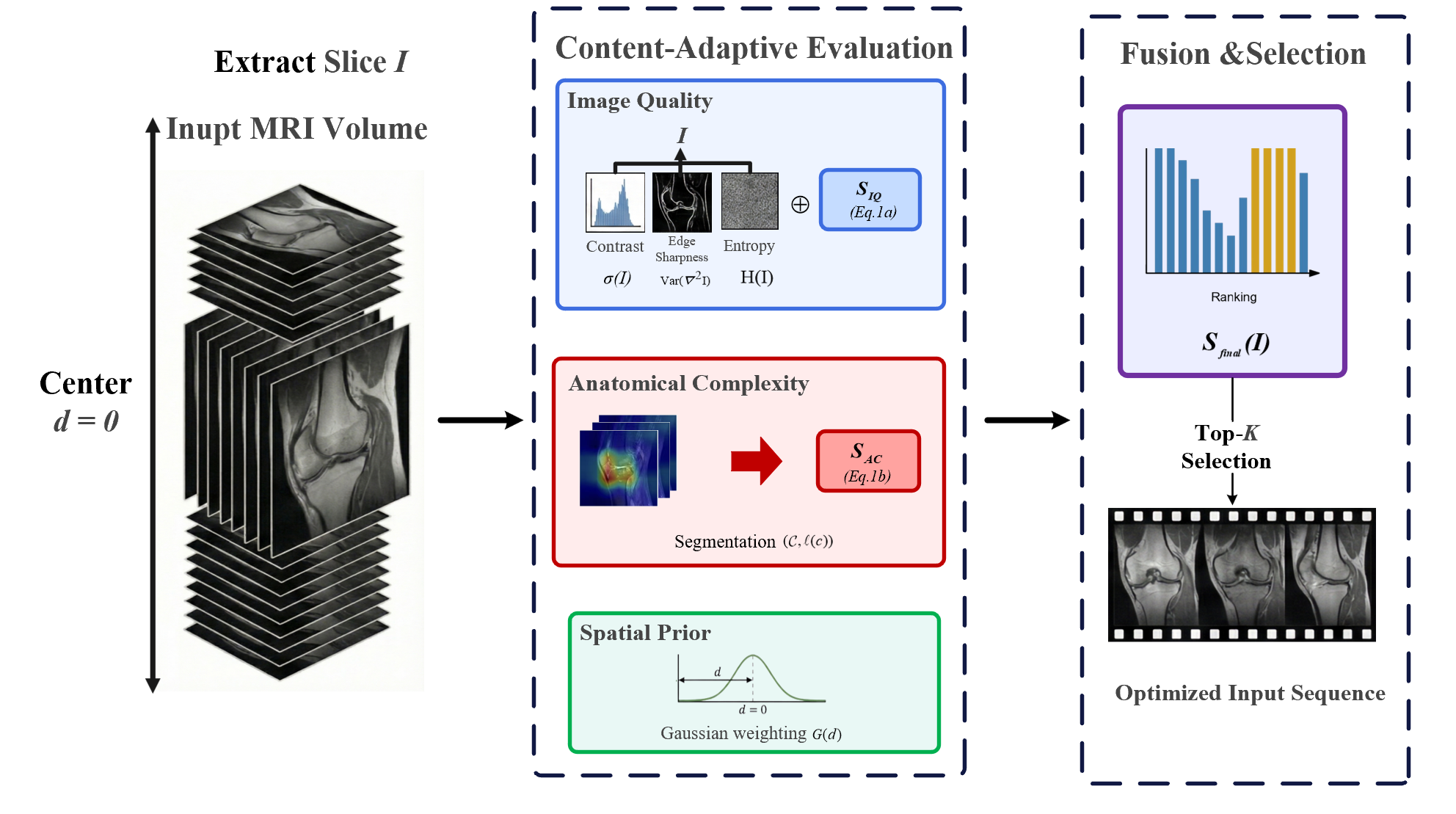} 
   \caption{Schematic illustration of the proposed content-adaptive MRI slice selection and optimization process.}
   \label{fig2}
   \vspace{-0.5cm}
\end{figure}
To define the final selection score $S_{\text{final}}$, we explicitly combine these metrics with a spatial prior. As meniscal tears are centrally located, we apply a Gaussian spatial weighting $G(d)$~\cite{R18}, where $d$ is the slice distance from the volume center. The unified formulation is defined as:

\begin{equation}
\label{eq:final_score}
S_{\text{final}}(I) = G(d) \cdot \left( \alpha S_{\text{IQ}} + (1-\alpha) S_{\text{AC}} \right)
\end{equation}

\noindent where $\alpha$ controls the trade-off between image quality and anatomical relevance. We set $K{=}16$ slices per view and $\alpha{=}0.6$. SAM-Med2D (ViT-B) serves only to compute $S_{\text{AC}}$; $S_{\text{IQ}}$ and $G(d)$ ensure robust ranking when SAM-Med2D confidence is low. As visualized in Figure~\ref{fig2}, the top-$K$ slices with the highest $S_{\text{final}}$ are extracted and concatenated into a video sequence, constituting the optimized input for the visual encoder.

\subsection{Non-Visual Modalities: Clinical Priors and Clinical Text}

While visual data provides the morphological basis for diagnosis, clinical reasoning relies heavily on patient context and semantic interpretation. MeniOmni integrates these non-visual modalities to support holistic assessment:

\subsubsection{Structured Clinical Priors}
The health of the meniscus is inextricably linked to biomechanics and patient demographics. We explicitly structure key metadata for each case—age, sex, and BMI—to provide clinically relevant physiological constraints that facilitate model learning. These tabular priors are essential for the Severity Grading task, enabling the model to calibrate degeneration risks and mitigate spurious visual correlations.

\subsubsection{Clinical Text Annotations}
For the Report Generation task, the radiologists annotated each case with paired structured texts, providing dense semantic supervision beyond simple classification labels:
\begin{itemize}
    \item \textbf{Morphological Description:} Detailed findings describing signal intensity and shape anomalies (e.g., \textit{``Linear high signal intensity in the posterior horn not extending to the articular surface.''}).
    \item \textbf{Diagnostic Conclusion:} The final clinical assessment derived from the findings (e.g., \textit{``Grade 2 injury to the lateral meniscus.''}).
\end{itemize}

\section{Proposed Benchmarking Framework}

We present the MeniOmni benchmarking framework, tailored to address the unique challenges of holistic meniscal assessment: the heterogeneity of multimodal inputs and the necessity for compartment-specific anatomical diagnosis.

\subsection{Benchmarking Pipeline Setup}

Clinical assessment of the knee necessitates the independent evaluation of the Medial and Lateral menisci, as they often exhibit divergent pathologies. Consequently, we formulate the severity grading task as a Dual-Compartment Prediction problem. The goal is to map the multimodal inputs—tri-planar MRI sequences $\mathcal{V}$ and tabular Clinical Priors $\mathcal{T}$—to a pair of compartment-wise Stoller grades $(y_{med}, y_{lat})$, where $y \in \{0, 1, 2, 3\}$.

\subsubsection{Clinical Prior Serialization}

Rather than employing specialized tabular encoders, we define a serialization function $\Phi: \mathcal{T} \rightarrow \mathcal{L}$ that transforms Clinical Priors (Age, Sex, BMI) into a linguistic embedding space. A patient record is serialized into a natural language prompt $T_{prior}$:

\begin{equation}
\label{eq2}
T_{prior} = \text{``A } t_{age} \text{-year-old } t_{sex} \text{ patient with a BMI of } t_{bmi}".
\end{equation}

This unification enables the model to interpret risk factors as semantic context rather than isolated numerical features.

\subsubsection{Tri-Stream Visual Perception}

To preserve the anatomical geometry of orthogonal MRI planes, we propose a Siamese Multi-Stream Encoding strategy. Let $\mathcal{V} = \{V_{sag}, V_{cor}, V_{axi}\}$ denote the input volumetric slice sequences. We employ a weight-shared Video Encoder $\mathcal{E}_{vid}$ to process each view independently, interpreting the $z$-axis depth as a pseudo-temporal dimension:

\begin{equation}
\label{eq3}
\mathbf{H}_{p} = \mathcal{E}_{vid}(V_{p}), \quad \forall p \in \{sag, cor, axi\},
\end{equation}
where $\mathbf{H}_{p} \in \mathbb{R}^{T \times D}$ represents the spatiotemporal tokens. To facilitate explicit anatomical correlation, we construct an interleaved visual sequence $\mathbf{X}_{visual}$ with view-specific textual identifiers:

\begin{equation}
\label{eq4}
\begin{aligned}
\mathbf{X}_{visual} = [ \quad &\texttt{<VID>}, \mathbf{H}_{sag}, \texttt{</VID>}, \text{\textit{``Sagittal View''}}, \\
&\texttt{<VID>}, \mathbf{H}_{cor}, \texttt{</VID>}, \text{\textit{``Coronal View''}}, \\
&\texttt{<VID>}, \mathbf{H}_{axi}, \texttt{</VID>}, \text{\textit{``Axial View''}} \quad ].
\end{aligned}
\end{equation}

\noindent where \texttt{<VID>} and \texttt{</VID>} denote special modality tokens that demarcate the boundaries of visual embeddings. The utilisation of these tokens facilitates the LLM's capacity to discern continuous visual representations from discrete textual tokens. This process effectively employs the LLM's positional encoding, thereby binding anatomical semantics with their respective visual features.

\subsection{Task Formulation}

\noindent\textbf{Task 1: Fine-Grained Severity Grading.}
This task emulates holistic clinical decision-making. We construct the input $\mathbf{X}_{I}$ by concatenating the visual stream with serialized priors:
\begin{equation}
\label{eq5}
\mathbf{X}_{I} = [\mathbf{X}_{visual}, T_{prior}, \mathbf{I}_{grade}].
\end{equation}
Including $T_{prior}$ allows the model to calibrate pre-test probabilities using physiological risks. We use a Structured Instruction $\mathbf{I}_{grade}$:\textit{{{``Analyze the multi-view MRI sequences and the patient's background information. Determine the Stoller severity grade (0-3) for both the Medial Meniscus (MM) and Lateral Meniscus (LM)."}}} to direct the model to output structured predictions.

\noindent\textbf{Task 2: Visual Report Generation.}
To evaluate the translation of pixel-level information into semantic findings, we restrict the input strictly to the visual stream to mitigate reliance on demographic shortcuts:
\begin{equation}
\label{eq6}
\mathbf{X}_{II} = [\mathbf{X}_{visual}, \mathbf{I}_{gen}].
\end{equation}
Using a Role-Playing Prompt $\mathbf{I}_{gen}$:\textit{{{``Serve as a senior radiologist. Focus strictly on the visual evidence. Describe the imaging features observed in the MRI sequences, specifically detailing signal intensity, morphology, and tear patterns. Conclude with a final impression."}}} , we guide the model to generate grounded radiological descriptions rather than unsubstantiated hallucinations.

\section{Task-Specific Evaluation Protocols}

Standard metrics often treat all errors uniformly or focus solely on linguistic fluency. To ensure clinical relevance, we propose two domain-specific protocols.

\subsection{Protocol I: Risk-Aware Ordinal Evaluation}

The conventional classification metrics (e.g., accuracy, F1-score) are inadequate for meniscal grading due to their assumption of categorical independence, which fails to account for the ordinal progression of degeneration ($0\to 1\to 2\to 3$). From a clinical perspective, the magnitude of error dictates patient safety. The radiological distinction between Grade 1 and 2 is of minor significance, as both conditions are typically managed with a conservative approach. However, misclassifying Grade 0 (Intact) as Grade 3 (Complete Tear) or vice versa constitutes a catastrophic failure that has the potential to result in the need for unnecessary arthroscopy or the failure to capitalise on surgical opportunities.

Generic metrics inadequately equate these scenarios, potentially validating models that are statistically performant but clinically reckless. In order to align evaluation with clinical risk stratification, a composite protocol is proposed, balancing correlation and safety:

\noindent\textbf{1. Ordinal Consistency: Average Quadratic Weighted Kappa (Avg-QWK).}We employ QWK as the primary metric to capture the semantic proximity between grades. Unlike linear metrics, QWK imposes a non-linear penalty $w_{i,j} = (i-j)^2 / (C-1)^2$ on disagreements. This quadratically penalizes substantial deviations while tolerating minor discrepancies, aligning with the continuous nature of cartilage degeneration. We compute the macro-average of medial and lateral compartments to mitigate class imbalance:

\begin{equation}
\label{eq7}
\text{Avg-QWK} = \frac{1}{2} (\kappa_{med} + \kappa_{lat}).
\end{equation}

\noindent\textbf{2. Clinical Safety: Global Severe Error Rate (SER).}While QWK aggregates average performance, it may mask critical misclassifications. SER acts as a hard safety constraint to explicitly track outliers. We define a "Severe Error" as a deviation of $\ge 2$ grades. The metric enforces a strict case-level tolerance: a diagnosis is deemed unsafe if either compartment exceeds the error threshold:

\begin{equation}
\label{eq8}
\text{SER} = \frac{1}{N} \sum_{k=1}^N \mathbb{I}\left( \max(|\Delta y_{med}^{(k)}|, |\Delta y_{lat}^{(k)}|) \ge 2 \right),
\end{equation}

where $\mathbb{I}(\cdot)$ is the indicator function. Lower SER implies the model makes fewer clinically hazardous errors.

\subsection{Protocol II: The Meni-Score (Semantic Factual Consistency)}

Standard n-gram metrics quantify surface-level lexical overlap but fail to capture factual consistency. In medical reporting, a "horizontal tear" and a "vertical tear" have high n-gram overlap but represent contradictory diagnoses. To quantify semantic precision, we introduce the Meni-Score, an entity-centric metric grounded in a structured clinical ontology.

\noindent\textbf{1. Structured Clinical Vocabulary Construction.}
Diverging from unstructured text matching, we define a discrete pathological vocabulary set $\mathcal{V}_{meni} = \mathcal{V}_{loc} \cup \mathcal{V}_{morph} \cup \mathcal{V}_{sig}$, comprising three disjoint semantic subsets distilled from radiological standards:

\begin{itemize}
\item \textbf{Location Subspace ($\mathcal{V}_{loc}$):} {\textit{"posterior horn", "anterior root", "body"}}.
\item \textbf{Morphology Subspace ($\mathcal{V}_{morph}$):} {\textit{"horizontal", "radial", "bucket-handle", "complex"}}.
\item \textbf{Signal Subspace ($\mathcal{V}_{sig}$):} {\textit{"linear high signal", "globular", "maceration"}}.
\end{itemize}

We implement a Canonicalization Function $\Phi_{\text{norm}}(\cdot)$ that maps synonymous variations to unique tokens in $\mathcal{V}_{\text{meni}}$, ensuring robustness against linguistic variations. For example, $\Phi_{\text{norm}}$ maps ``high signal'' $\to$ ``linear high signal'' and ``meniscal tear'' $\to$ the corresponding morphology token in $\mathcal{V}_{\text{morph}}$. The complete entity extraction rules, synonym mappings, and clinical ontology will be released as part of our open source evaluation toolkit.

\noindent\textbf{2. Set-Theoretic Calculation.}
Let $R_{gen}$ be the generated report and $R_{gt}$ be the ground truth. We employ an entity extractor $Extract(.)$ to map both texts into the entity space: $S_{gen} = \Phi_{norm}(Extract(R_{gen}))$ and $S_{gt} = \Phi_{norm}(Extract(R_{gt}))$, where $S \subset \mathcal{V}_{meni}$.
The Meni-Score is formulated as the F1-score of the clinical precision and recall based on set intersection:
\begin{equation}
\label{eq:meniscore}
\text{Meni-Score} = 2 \cdot \frac{P_{clin} \cdot R_{clin}}{P_{clin} + R_{clin}},
\end{equation}
where $P_{clin} = \frac{|S_{gen} \cap S_{gt}|}{|S_{gen}|}$ serves as a proxy for hallucination control, and $R_{clin} = \frac{|S_{gen} \cap S_{gt}|}{|S_{gt}|}$ measures the information coverage.

\section{Experimental Results Analysis}

We evaluate the proposed benchmarks across two distinct tasks: Severity Grading (Task 1) and Report Generation (Task 2). This section analyzes the quantitative performance of supervised visual specialists and state-of-the-art MLLMs.

\subsection{Task 1: Fine-Grained Severity Grading}
Table~\ref{tab:task1} presents the comparison on meniscal severity grading. We compare supervised specialists, which are fine-tuned on the training set to learn a direct mapping from voxels to labels, against generalist MLLMs evaluated in a zero-shot setting using the serialized prior $T$ and interleaved visual sequence $X_{\text{visual}}$. Zero-shot MLLMs use a temperature of $0$ for grading (deterministic) and $0.3$ for report generation (controlled diversity).

\begin{table}[h]
\centering
\vspace{-0.25cm}
\caption{Task 1 Performance: Meniscal Severity Grading.}
\label{tab:task1}
\small
\setlength{\tabcolsep}{3.5pt}
\begin{tabular}{l cccc}
\toprule
\textbf{Model} & \textbf{Acc} & \textbf{F1} & \textbf{QWK} & \textbf{SER} $\downarrow$ \\
\midrule
\multicolumn{5}{l}{\textit{Supervised Specialists (Visual-Only, Fine-tuned)}} \\
Inception-3D & 56.5\% & 53.8\% & 0.495 & 16.2\% \\
ResNet3D-50 & 58.7\% & 56.3\% & 0.525 & 14.5\% \\
X3D-L\cite{R24} & 61.2\% & 58.9\% & 0.558 & 12.8\% \\
ViViT-B\cite{R23} & 62.9\% & 60.5\% & 0.592 & 11.0\% \\
\textbf{Video Swin-B}\cite{R22} & \textbf{63.8\%} & \textbf{61.2\%} & \textbf{0.605} & \textbf{10.2\%} \\
\midrule
\multicolumn{5}{l}{\textit{Open-Source MLLMs (Zero-shot)}} \\
Video-LLaVA~\cite{R19} & 38.5\% & 34.2\% & 0.215 & 35.4\% \\
InternVL2~\cite{R20} & 51.2\% & 48.1\% & 0.448 & 20.5\% \\
Qwen2-VL-72B~\cite{R21} & 55.4\% & 52.6\% & 0.492 & 17.8\% \\
\midrule
\multicolumn{5}{l}{\textit{Proprietary MLLMs (Zero-shot)}} \\
Gemini 1.5 Pro & 58.4\% & 55.1\% & 0.518 & 15.2\% \\
Claude 3.5 Sonnet & 60.2\% & 57.5\% & 0.562 & 13.8\% \\
\textbf{GPT-4o} & \textbf{62.5\%} & \textbf{59.8\%} & \textbf{0.585} & \textbf{12.4\%} \\
\bottomrule
\end{tabular}
   \vspace{-0.25cm}
\end{table}

As hypothesised, supervised specialists currently dominate the leaderboard, with the Video Swin-B model achieving state-of-the-art accuracy (63.8\%) and QWK (0.605). This validates the efficacy of hierarchical attention mechanisms in modelling the anisotropic 3D nature of meniscal tears and in capturing fine-grained morphological discontinuities that are often missed by generic encoders. However, GPT-4o's performance is also impressive, with an accuracy of 62.5\%, trailing by only 1.3\% in a zero-shot setting. This implicitly validates our Tri-Stream Visual Perception and Clinical Prior Serialisation strategies, demonstrating that MLLMs can successfully perform 'cognitive registration' of orthogonal planes and leverage Clinical Priors to calibrate diagnostic probabilities. Nevertheless, a critical divergence remains in terms of reliability: while specialists maintain a robust severe error rate (SER) of 10.2\% due to their constrained classification manifold, open-source MLLMs exhibit prohibitive volatility (e.g. Video-LLaVA at 35.4\%). This confirms that, although generalist MLLMs have high potential, they currently lack the robustness of fine-tuned specialists, highlighting the importance of the SER metric in identifying catastrophic clinical errors.

\subsection{Task 2: Visual Report Generation}

Table~\ref{tab:task2} assesses the ability to translate pixel-level evidence into well-grounded radiological reports. This task is unique to MLLMs.

\begin{table}[h]
\centering
\caption{Task 2 Performance: Diagnostic Report Generation.}
\label{tab:task2}
\small
\setlength{\tabcolsep}{5pt}
\begin{tabular}{l ccc}
\toprule
\textbf{Model} & \textbf{BLEU-4} & \textbf{ROUGE-L} & \textbf{Meni-Score} \\
\midrule
\multicolumn{4}{l}{\textit{Open-Source MLLMs}} \\
Video-LLaVA~\cite{R19} & 15.2 & 28.5 & 32.4 \\
InternVL2~\cite{R20} & 20.1 & 36.5 & 44.2 \\
Qwen2-VL-72B~\cite{R21} & 22.5 & 39.8 & 46.5 \\
\midrule
\multicolumn{4}{l}{\textit{Proprietary MLLMs}} \\
Gemini 1.5 Pro & 24.5 & 41.2 & 48.5 \\
Claude 3.5 Sonnet & \textbf{27.2} & 43.8 & 51.0 \\
\textbf{GPT-4o} & 26.8 & \textbf{44.5} & \textbf{52.4} \\
\bottomrule
\end{tabular}
   \vspace{-0.4cm}
\end{table}

Our evaluation reveals a steep performance scaling law in diagnostic report generation. The ability to interpret complex visual features and translate them into precise medical terminology is a distinct property of model scale. While smaller models such as Video-LLaVA struggle with entity recall, large-scale foundation models such as Qwen2-VL-72B and GPT-4o demonstrate significantly higher alignment with the clinical vocabulary $\mathcal{V}_{meni}$. Crucially, however, the results reveal a discrepancy between linguistic fluency and clinical accuracy: although Claude 3.5 Sonnet achieves the highest BLEU-4 score (27.2) thanks to its superior fluency, GPT-4o surpasses it with a Meni-Score of 52.4. This discrepancy highlights the limitations of standard n-gram metrics and confirms that top-tier proprietary models excel specifically in semantically factual consistency, minimising radiological hallucinations, as verified by our rigorous, ontology-grounded benchmarking protocol.

\section{Ablation Study: Effectiveness of Clinical Priors}
A core hypothesis of MeniOmni is that accurate diagnosis requires holistic patient profiling. To validate this, we conduct an inference-time ablation with GPT-4o, comparing a (A) Vision Only baseline against our (B) Multimodal formulation (Vision + Age/Sex/BMI).

\noindent\textbf{Experimental Settings.}
\begin{itemize}
    \item \textbf{(A) Vision Only:} The model receives only the tri-stream video sequence without any textual metadata.
    \item \textbf{(B) Vision + Clinical Priors (Ours):} We integrate the visual input with basic patient demographics and clinical indicators (Age, Sex, BMI).
\end{itemize}

\begin{table}[h]
\centering
\small
\caption{Ablation Study on Clinical Priors.}
\label{tab:ablation_priors}
\begin{tabular*}{\columnwidth}{l @{\extracolsep{\fill}} c c c}
\toprule
\multirow{2}{*}{\textbf{Input Configuration}} & \multicolumn{2}{c}{\textbf{Consistency} ($\uparrow$)} & \textbf{Safety} ($\downarrow$) \\
\cmidrule(lr){2-3} \cmidrule(lr){4-4}
 & \textbf{Acc} & \textbf{Avg-QWK} & \textbf{SER} \\
\midrule
(A) Vision Only & 58.4\% & 0.512 & 16.8\% \\
\textbf{(B) Vision + Priors} & \textbf{62.5\%} & \textbf{0.585} & \textbf{12.4\%} \\
\bottomrule
\end{tabular*}
   \vspace{-0.25cm}
\end{table}

\noindent\textbf{Demographics as Bayesian Filters.}
As shown in Table~\ref{tab:ablation_priors}, integrating clinical priors substantially improves performance, increasing accuracy by 4.1\% and average QWK by 0.073. This confirms that demographics function as effective 'Bayesian priors', enabling the model to calibrate decision boundaries based on epidemiological norms. For example, the model learns to reduce the probability assigned to degenerative tears in younger patients, even when visual signals are ambiguous.

\begin{figure}[t]
    \centering
    \includegraphics[width=\columnwidth]{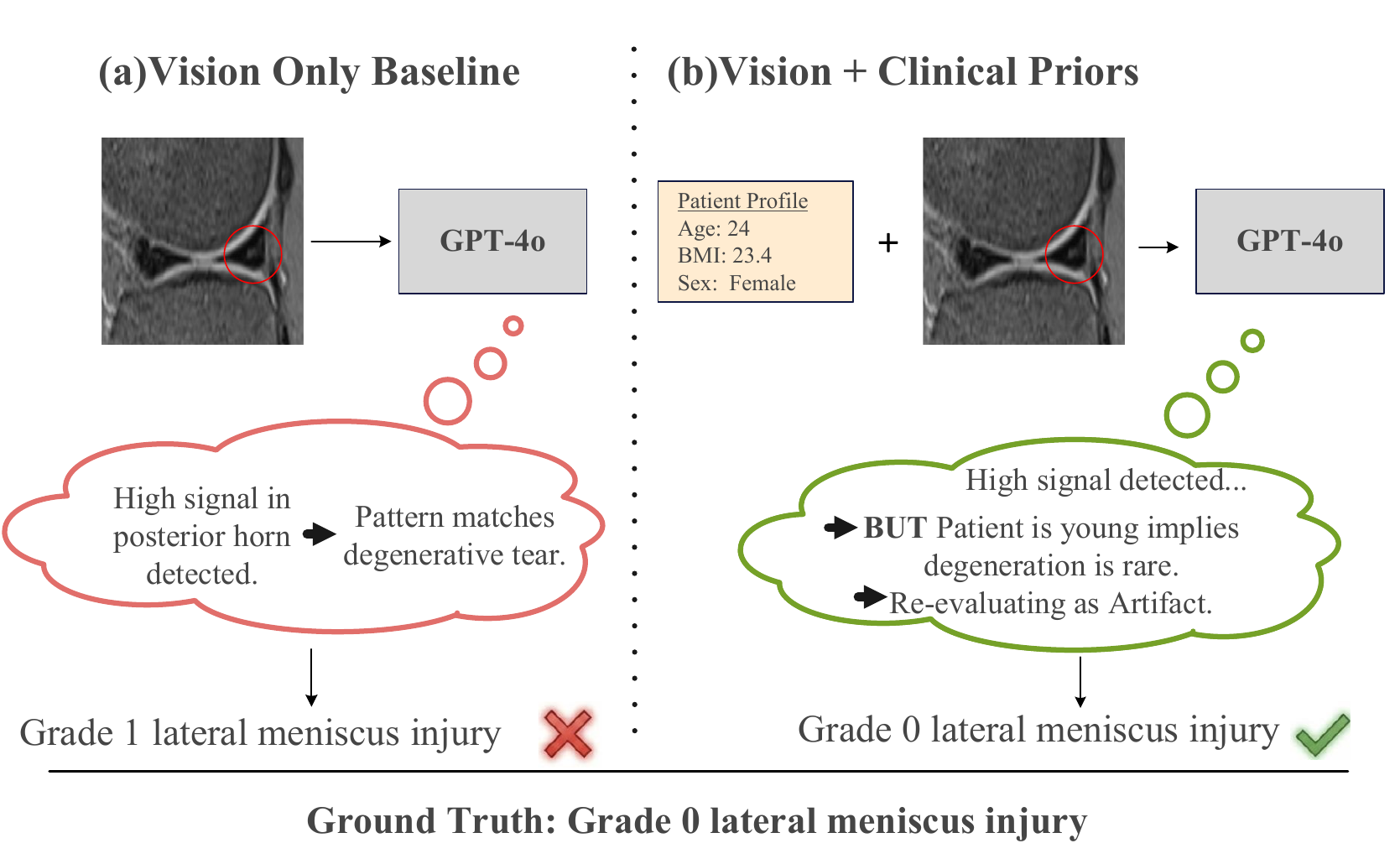}
    \caption{Qualitative Analysis. The integration of clinical priors (Age: 26) corrects a visual hallucination, downgrading a false Grade 2 diagnosis to Grade 0.}
    \label{fig:qualitative_case}
    \vspace{-0.4cm}
\end{figure}

\noindent\textbf{Safety Mechanism and Qualitative Correction.}
Crucially, priors act as a safety regulariser, causing a significant decrease in the Severe Error Rate (SER) from 16.8\% to 12.4\%. As shown in Figure~\ref{fig:qualitative_case}, the Vision-Only model falls into a 'Visual Ambiguity Trap', misinterpreting a hyperintense signal in the posterior horn as a degenerative tear (Grade 2). However, when the prior 'Age: 26' is integrated, the Multimodal model performs a Bayesian update. Recognising that primary degeneration is uncommon in young adults, it suppresses the visual hallucination and correctly reclassifies the signal as a benign artefact (grade 0). This demonstrates how MeniOmni aligns diagnostic reasoning with clinical plausibility, effectively filtering out catastrophic false positives.

\section{Conclusion}

We presented MeniOmni, a structured multimodal benchmark shifting meniscus injury assessment from isolated visual perception to holistic clinical reasoning by integrating multi-view MRI, Clinical Priors, and diagnostic text. Our benchmarking reveals that while proprietary models outperform open-source alternatives, they remain susceptible to hallucinations with visual-only inputs. The integration of clinical priors functions as an effective Bayesian filter, substantially reducing the Severe Error Rate. MeniOmni serves as both a comprehensive testbed and a step towards safer, clinically aligned AI for orthopedic diagnostics.

\bibliographystyle{IEEEbib}
\bibliography{icme2026references}

\end{document}